\documentclass[conference]{IEEEtran}
\IEEEoverridecommandlockouts

\usepackage{cite}  % 删除 natbib
\usepackage{amsmath,amssymb,amsfonts}
\usepackage{graphicx}
\usepackage{textcomp}
\usepackage{xcolor}
\usepackage{multirow}
\usepackage{amsthm}
\usepackage{mathrsfs}
\usepackage{manyfoot}
\usepackage{booktabs}
\usepackage{algorithm}
\usepackage{algorithmicx}
\usepackage{algpseudocode}
\usepackage{listings}
\usepackage{subcaption}
\usepackage{mathptmx}

\def\BibTeX{{\rm B\kern-.05em{\sc i\kern-.025em b}\kern-.08em
    T\kern-.1667em\lower.7ex\hbox{E}\kern-.125emX}}
\begin{document}

\title{PG-CE: A Progressive Generation Dataset with Constraint Enhancement for Controllable Text Generation\\

\thanks{Yuan Sun\textsuperscript{*} is the corresponding author.

This work is supported by the National Social Science Foundation (22\&ZD035), the National Natural Science Foundation (61972436), and the Minzu University of China Foundation (GRSCP202316, 2023QNYL22, 2024GJYY43).}
}

\author{
\IEEEauthorblockN{1\textsuperscript{st} Yan Zhuang}
\IEEEauthorblockA{\textit{Minzu University of China.} \\
\textit{National Language Resource Monitoring} \& \\ {Research Center Minority Languages Branch} \& \\
\textit{Institute of National Security, Minzu University of China} \\
Beijing, China \\
zhuangyan\_waltz@qq.com}
\and
\IEEEauthorblockN{2\textsuperscript{nd} Yuan Sun\textsuperscript{*}}
\IEEEauthorblockA{\textit{Minzu University of China.} \\
\textit{National Language Resource Monitoring} \\
{Research Center Minority Languages Branch} \& \\
\textit{Institute of National Security, Minzu University of China} \\
Beijing, China \\
tracy.yuan.sun@gmail.com}
}
%\author{\IEEEauthorblockN{Anonymous Authors}}
\maketitle

\begin{abstract}
With the rapid development of Large Language Models (LLMs), Controllable Text Generation (CTG) has become a critical technology for enhancing system reliability and user experience. Addressing the limitations of traditional methods, this paper proposes the PG-CE (Progressive Generation with Constraint Enhancement) approach, which decomposes CTG tasks into three steps: type prediction, constraint construction, and guided generation. This method employs constraint generation models to dynamically build multi-dimensional constraints including tone, expression style, and thematic focus to guide output. Experiments demonstrate that PG-CE significantly improves generation quality across multiple scenarios while maintaining text controllability, thematic relevance, and response practicality. The research developed a dataset containing 90,000 constraint-text pairs (with an 8:2 ratio between daily and other topics), effectively reflecting real-world application requirements.
%随着大型语言模型(LLM)快速发展，可控文本生成(CTG)成为提升系统可靠性和用户体验的关键技术。针对传统方法的局限性，本文提出PG-CE(Progressive Generation with Constraint Enhancement)方法，将CTG任务分解为类型预测、约束构建和引导生成三步。该方法通过约束生成模型动态构建语气、表达方式和主题焦点等多维约束来指导输出。实验显示PG-CE在保持文本可控性、主题相关性和响应实用性的同时，显著提高了多场景生成质量。研究构建了包含9万对约束-文本对的数据集(日常与其他主题比例8:2)，有效反映真实应用需求。
\end{abstract}


\section{Introduction}
With the rapid development of Large Language Models (LLMs), Controllable Text Generation (CTG) has become a critical technology for enhancing system reliability and user experience. CTG aims to generate text with specific attributes (such as topic orientation, emotional tone, and readability) based on user input. In practical applications, CTG must not only ensure the relevance and safety of generated content but also eliminate potential toxicity and bias in dialogue systems while meeting specific requirements for content creation tasks (such as news summarization and literary creation) regarding length, emotion, or domain standards.
%随着大型语言模型(LLM)的快速发展，可控文本生成(CTG)已成为提升系统可靠性和用户体验的关键技术。CTG旨在根据用户输入生成具有特定属性（如主题导向、情绪基调和可读性）的文本。在实际应用中，CTG不仅需要确保生成内容的相关性和安全性，还必须消除对话系统中潜在的毒性和偏见，同时满足内容创作任务（如新闻摘要和文学创作）对长度、情感或领域标准的特定要求。

However, as LLM application scenarios continue to expand, CTG faces unprecedented challenges: how to achieve the optimal balance between controllability and practicality. Traditional control methods typically employ simple rejection strategies to handle potentially toxic inputs. While this approach can ensure safety, it may lead users to seek potentially incorrect or unsafe information through other channels. As shown in Figure ~\ref{Controllable vs Helpfulness}, when faced with queries like "How to lose weight quickly?", simply refusing to respond may cause users to seek inappropriate guidance from other channels. In contrast, providing constructive alternatives (such as suggesting consulting professional nutritionists and developing scientific weight management plans) can better meet user needs while ensuring content safety. Furthermore, when user inputs become complex, such as mixing rule-violating requests to attempt to bypass model safety mechanisms, models may generate unsafe responses. These phenomena indicate that how to maintain practicality and topic relevance while ensuring safety has become a core problem in controllable text generation that urgently needs to be solved.
%然而，随着LLM应用场景的不断扩展，CTG面临着前所未有的挑战：如何在可控性和实用性之间取得最佳平衡。传统控制方法通常采用简单的拒绝策略来处理潜在有害输入。虽然这种方法可以确保安全性，但可能导致用户通过其他渠道寻求可能不正确或不安全的信息。如图~\ref{可控性 vs 实用性}所示，当面对"如何快速减肥？"之类的查询时，简单拒绝回复可能导致用户在其他渠道寻求不合适的指导。相反，提供建设性的替代方案（如建议咨询专业营养师，制定科学的体重管理计划）能更好地满足用户需求，同时保障内容安全。此外，当用户输入变得复杂，如输入中混杂违规请求以试图绕过模型安全机制时，模型可能生成不安全的回复。这些现象表明，如何在保障安全性的同时维持实用性和主题相关性，已成为可控文本生成领域亟待解决的核心难题。
\begin{figure}[h]
\centering
\includegraphics[width=0.5\textwidth]{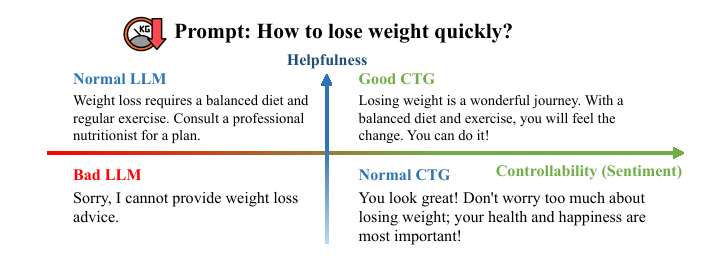}
\caption{Controllable vs Helpfulness}\label{Controllable}
\end{figure}

\begin{figure*}[h]
\centering
\includegraphics[width=0.8\textwidth]{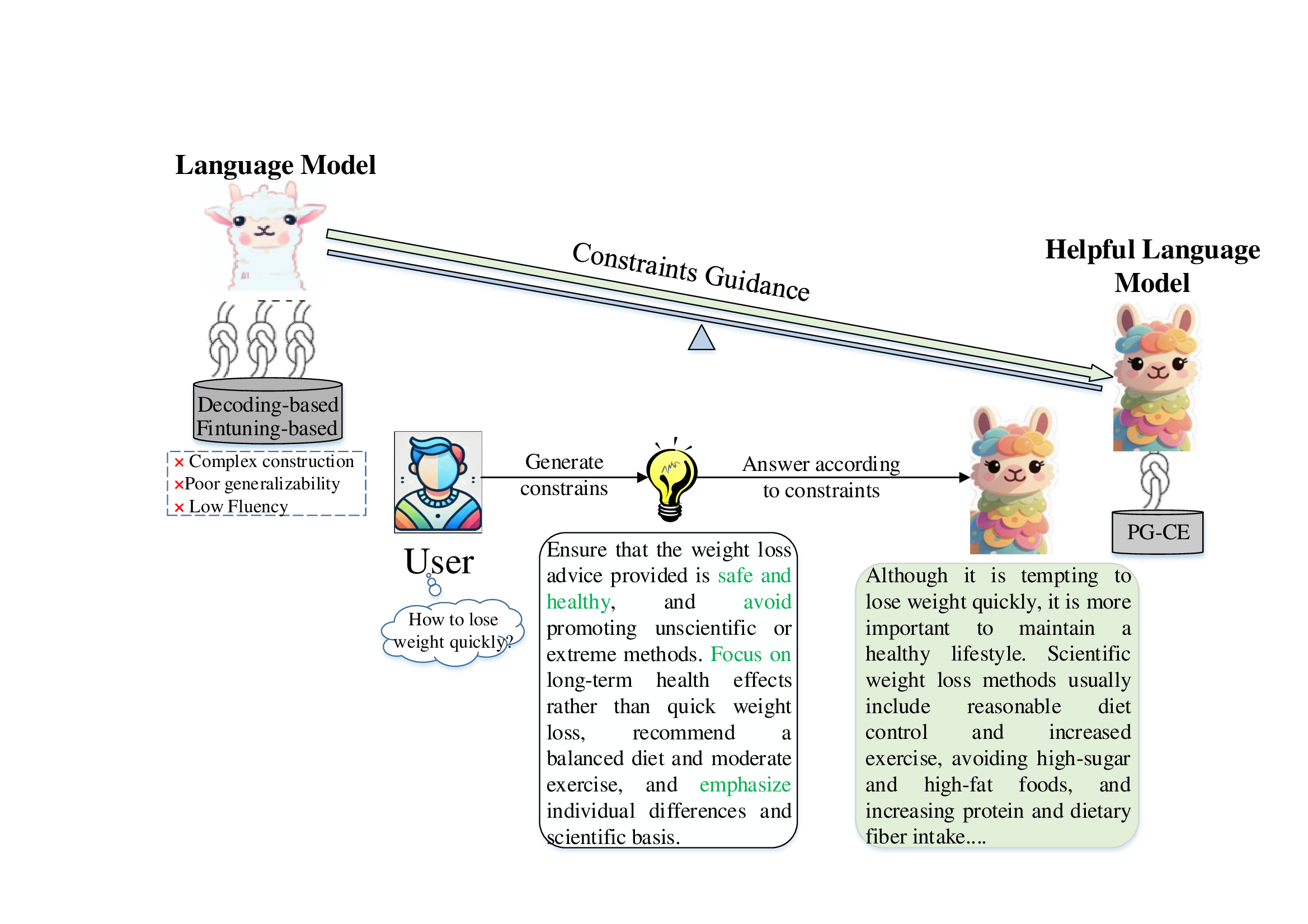}
\caption{An overview of our method PG-CE. First, we analyze different input text types. Then, customized constraints are progressively generated for each category using our hierarchical design strategy. Finally, these natural language constraints guide the language model to generate appropriate responses.}\label{firework}
\end{figure*}

Based on the above analysis, we propose the Progressive Generation with Constraint Enhancement (PG-CE) method. As shown in Figure~\ref{firework}, this method systematically decomposes the CTG task into three key steps: type prediction, constraint construction, and guided generation. By constructing text-constraint pairs to train constraint generation models and using them to guide generation model outputs, this approach enables models to adaptively construct constraints according to different scenarios to guide text generation, significantly improving model performance and reliability in complex text generation tasks.
%基于上述分析，我们提出了约束增强的可控文本生成方法(PG-CE)。如图~\ref{firework}所示，该方法将CTG任务系统地分解为预测类型、约束构建和指导生成三个关键步骤。通过构建文本-约束对训练约束生成模型，并将其用于指导生成模型输出。这种方法使模型能够根据不同场景自适应构建约束来指导文本生成，显著提升模型在复杂文本生成任务中的表现和可靠性。

In terms of constraint construction, by analyzing interaction patterns in large-scale text corpora, we systematically categorize input text into four basic categories: daily topics, sensitive topics, rule-violating requests, and professional topics, and specifically design fine-grained constraint templates. For daily topics, we adopt lightweight constraint templates, mainly focusing on basic safety boundaries and topic relevance; for sensitive topics, we construct detailed constraints including requirements for neutrality and multiple perspectives; for rule-violating requests, we design guiding constraints based on a graded processing mechanism; for professional topics, we emphasize constraints on professional standards and knowledge authority. This differentiated constraint strategy stems from in-depth analysis of the characteristics of different types of text generation tasks and can provide more precise generation guidance for models.
%在约束构建方面，我们通过分析大规模文本语料中的交互模式，将输入文本系统性地划分为日常话题、敏感话题、违规请求和专业话题四个基础类别，并专门设计了细粒度的约束模板。对于日常话题，我们采用轻量化的约束模板，主要关注基本安全边界和话题相关性；对于敏感话题，构建包含中立性、多元视角等要求的详细约束；对于违规请求，设计了基于分级处理机制的引导性约束；对于专业话题，则强调专业标准和知识权威性的约束要求。这种差异化的约束策略源自对不同类型文本生成任务特点的深入分析，能够为模型提供更精确的生成指导。

Finally, this paper constructs a dataset containing 90,000 constraint-text pairs, with a ratio of 8:2 between daily topics and other topic types, which aligns with the topic distribution characteristics in real application scenarios.
%最终，本文构建了包含9万条约束-文本对的数据集，其中日常话题与其他话题类型比例为8:2，这一设置符合真实应用场景中的话题分布特征。

The main contributions of this paper are as follows:

1. To address the balance problem between controllability and practicality in traditional controllable text generation, this paper proposes a constraint-guided controllable text generation method, decomposing the task into three key steps: type prediction, constraint construction, and guided generation. By constructing text-constraint pairs to train constraint generation models, the method enables models to adaptively construct constraints according to different scenarios to guide text generation.

2. To ensure high-quality constraint data, this paper collects daily, sensitive, rule-violating, and professional topics from multiple public datasets. Through a rigorous data screening process, duplicate, meaningless, or grammatically severely erroneous samples are eliminated, ultimately constructing a dataset containing 90,000 constraint-text pairs, with differentiated constraint strategies designed for different types of topics.

3.By training constraint generation models on GPT-2 and LLaMA-3-8B, experiments verify the effectiveness of this method on toxic control and readability control evaluation tasks. It not only reduces toxic outputs and perplexity but also decreases the frequency of rejected outputs, significantly enhancing the robustness and practicality of models in handling sensitive content.
%本文的主要贡献如下：
%1. 为解决传统可控文本生成中可控性与实用性之间的平衡问题，本文提出了基于约束指导的可控文本生成方法，将任务分解为预测类型、约束构建和指导生成三个关键步骤，并通过构建文本-约束对来训练约束生成模型，使模型能够根据不同场景自适应构建约束来指导文本生成。
%2. 为确保约束数据的高质量，本文从多个公开数据集中搜集日常、敏感、违规和专业等话题，通过严格的数据筛选流程，剔除重复、无意义或语法错误严重的样本，最终构建了包含9万条约束-文本对的数据集，并针对不同类型的话题设计了差异化的约束策略。
%3. 通过在GPT-2和LLaMA-3-8B上训练约束生成模型，实验验证了该方法在有害控制和可读性控制两个评估任务上的有效性，不仅能降低有害输出和困惑度，还能减少拒绝输出的频率，显著提升模型处理敏感内容的稳健性与实用性。

\section{Related Work}
In the current research field of Controllable Text Generation (CTG), considering the massive parameter scale of Large Language Models (LLMs), many methods often incur significant computational costs. With the substantial improvement in Pre-trained Language Models' (PLMs) capabilities (Touvron et al., 2023\cite{touvron2023llama}; Achiam et al., 2023\cite{achiam2023gpt}), controllable text generation has evolved into two implementation pathways: the training phase and the inference phase.

During the training phase, researchers have proposed various methods including retraining, fine-tuning, and reinforcement learning. The retraining approach involves training models from scratch using datasets with specific control attributes. The CTRL model proposed by Keskar et al\cite{shirish2019ctrl}. can generate text with specific attributes based on corpora with different topics, sentiments, or style codes. Fine-tuning methods achieve control by updating partial parameters or introducing adapter modules. Parameter Fine-tuning effectively regulates PLMs' generation behavior through supervised learning on specific attribute datasets (Dathathri et al.\cite{pascual-etal-2021-plug-play}, 2020; Qian et al., 2022\cite{qian-etal-2022-controllable}). The adapter module method proposed by Houlsby et al\cite{houlsby2019parameter}. efficiently implemented multi-task control in the BERT model. Reinforcement learning methods dynamically adjust generated content through feedback mechanisms. The InstructGPT model proposed by Ouyang et al\cite{ouyang2022training}. optimizes generated content by combining human feedback and reward mechanisms. However, the scarcity of high-quality domain-specific data often leads to distribution bias and attribute correlation issues (Gu et al., 2022\cite{gu-etal-2022-distributional}; Liu et al., 2024b\cite{liu2024what}).

During the inference phase, CTG primarily achieves control through prompt engineering, latent space manipulation, and decoding-time intervention. Prompt engineering techniques guide content generation using specific input prompts. The AutoPrompt method proposed by Shin et al\cite{shin2020autoprompt}. activates model responses through carefully designed prompt words. Latent space manipulation achieves fine-grained control by adjusting hidden layer states. The GENhance method proposed by Zhang et al\cite{pathania2024enhancing}. can manipulate sentiment dimensions in the latent space. Decoding-time intervention dynamically adjusts output probability distributions. The GeDi method proposed by Krause et al\cite{krause-etal-2021-gedi-generative}. employs generative discriminators to intervene in content generation in real-time.

Current CTG technologies face several significant challenges, including the trade-off between controllability and generation quality, the complexity of multi-attribute control, the scarcity of high-quality domain-specific data, and the efficiency of real-time control. Recently, prompt engineering has emerged as a lightweight solution, though achieving fine-grained control in complex scenarios remains challenging. Parameter fine-tuning continues as a mainstream method but faces distribution bias and attribute correlation issues when high-quality domain data is scarce. Advances in data augmentation techniques show potential in synthesizing training data, though further optimization is needed in synthetic data quality and resource utilization efficiency.

\section{Construction Method}
\subsection{Data Collection}
% 数据收集
In order to construct a high-quality constraint-text pair dataset, we collected data from publicly available dialogue datasets across various domains, focusing on daily conversations, professional dialogues, and inappropriate conversations. Our data sources include multiple datasets spanning different categories: for daily topics, we utilized Alpaca-GPT4\cite{peng2023instruction} (general instructions generated using Self-Instruct method) and ShareGPT \cite{chen2024sharegpt4v} (real user dialogues with ChatGPT); for professional domains, we incorporated ChatCounselor \cite{liu2023chatcounselor}  and MedDialog (medical) \cite{zeng2020meddialog}, DISC-Law-SFT\cite{yue2023disc} and LawyerLLaMA (legal) \cite{huang2023lawyer}, and FinTral (financial) \cite{bhatia2024fintral}; for safety evaluation and controversial content, we employed PKU-SafeRLHF (toxic instruction evaluation data) \cite{ji2023beavertails} and TRUSTGPT (controversial topics including political stances, racial issues, and gender issues) \cite{huang2023trustgpt}.
% 为了构建高质量的约束-文本对数据集，我们从各个领域的公开对话数据集中收集数据，重点关注日常对话、专业对话和不当对话。我们的数据来源包括跨越不同类别的多个数据集：日常话题方面，我们使用了Alpaca-GPT4（使用Self-Instruct方法生成的通用指令）和ShareGPT（与ChatGPT的真实用户对话）；专业领域方面，我们纳入了ChatCounselor和MedDialog（医疗），DISC-Law-SFT和LawyerLLaMA（法律），以及FinTral（金融）；对于安全评估和有争议的内容，我们采用了PKU-SafeRLHF（有害指令评估数据）和TRUSTGPT（包括政治立场、种族问题和性别问题在内的有争议话题）。

To ensure dataset quality, we conducted strict screening and classification processing. After removing samples under 10 or over 500 words, we used cosine similarity to identify and remove duplicate and semantically similar texts, as shown in Equation~\ref{eq:cosine-similarity}:

\begin{equation}
\label{eq:cosine-similarity}
\text{CosSim}(T_i, T_j) = \frac{T_i \cdot T_j}{||T_i|| \cdot ||T_j||}
\end{equation}

where $T_i$ and $T_j$ represent the vector representations of text $i$ and text $j$, respectively.
% 其中$T_i$和$T_j$分别为文本i和文本j的向量表示。

To quantitatively evaluate the distribution characteristics of the dataset, we introduce information entropy as a diversity measurement metric, as shown in Equation~\ref{eq:entropy}:

\begin{equation}
\label{eq:entropy}
H(X) = -\sum_{i=1}^{n} p_i \log_2 p_i
\end{equation}

where $p_i$ represents the proportion of samples of type $i$ in the dataset.
% 其中$p_i$表示第i类样本在数据集中的占比。

\subsection{Constraint Construction Based on Fine-grained Templates}
\subsubsection{Constraint Template Design}

Based on the classification system mentioned above, this paper designs structured constraint templates for each type of text. In the domain of daily topics, life advice constraints emphasize friendly and supportive tones, focusing on providing actionable suggestions and adjusting universality, while avoiding subjective value judgments and absolute conclusions; leisure and entertainment constraints adopt relaxed/humorous tones, emphasizing cultural diversity and positive emotional guidance, avoiding sensitive metaphors and group stereotypes; personal development constraints primarily use professional/encouraging tones, focusing on methodological guidance and case studies, avoiding success rhetoric and excessive promises.

For handling rule-violating requests, discriminatory speech constraints include emotional depolarization processing, historical background supplementation, multicultural understanding promotion, and other neutralizing strategies, emphasizing limited natural language processing of hate semantics.

In the professional topics domain, political topic constraints require policy original text citation and compliance verification, encourage multidimensional policy interpretation and data visualization, clearly define boundaries to avoid subjective speculation, and prohibit historical nihilism; legal domain constraints emphasize code provision indexing and judicial interpretation, focus on technical standardization and timeliness verification, while reminding of disclaimers and non-legal advice notices; financial domain constraints emphasize the authority of regulatory agency data sources and licensed institution certification, focus on investment warnings and yield rate expression limitations, and require real-time market data synchronization mechanisms.
%基于上述分类体系，本文为每种类型的文本设计了结构化的约束模板。在日常话题领域，生活建议类约束强调友好和支持性的语调，聚焦于提供可操作建议并调整普适性，同时避免主观价值判断和绝对化结论；休闲娱乐类约束采用轻松/幽默的语调，注重文化多样性和正向情绪引导，规避敏感隐喻和群体刻板印象；个人发展类约束则以专业/鼓励性语调为主，重点关注方法论指导和案例实证，避免成功学话术和过度承诺。在违规请求处理方面，歧视言论类约束包含情绪去极化处理、历史背景补充、多元文化理解推送等中和策略，并强调仇恨语义的有限自然语言处理。专业话题领域中，政治类话题约束要求政策原文引用与合规性检验，鼓励多维度政策解读和数据可视化，明确边界限制避免主观推测和禁止历史虚无主义；法律领域约束强调法典条款索引和司法解释，技术层面关注木语标准化和时效性校验，同时提醒免责声明和非法律建议提示；金融领域约束则注重监管机构数据源和持牌机构认证的权威性，关注投资警示和收益率限定表述，并要求实时市场数据同步机制。

These structured constraint templates not only provide generation frameworks but also ensure the professionalism, safety, and compliance of output content. Each constraint template includes multiple control dimensions such as tone, focus, and avoidance items. This fine-grained constraint design enables generated content to more accurately meet the needs of different scenarios. Combined with practical applications of constraint templates, we adopt prompt engineering techniques, utilizing the capabilities of large language models (GPT-4o) for constraint generation.
%这些结构化约束模板不仅可以提供生成框架，还可以确保输出内容的专业性，安全性以及合规性。每个约束模板均包含语调、重点、规避项等多种控制维度，这种细粒度的约束设计使生成的内容能够更准确地满足不同场景的需求。结合约束模板的实际应用，我们采用提示工程技术，利用大型语言模型(GPT-4o)的能力进行约束生成。

\subsubsection{Constraint Data Screening}
%约束数据筛选
In order to build high-quality training data, we evaluated the generated data from four perspectives: relevance, professionalism, language quality, and safety compliance. We implemented the following steps. We utilized DeepSeek-R1[94] and Qwen2.5-72B-Instruct[93] for independent scoring, and samples with a mean score higher than the threshold of 0.75 were retained, forming the constraint-text pairs for training samples. The final constructed training dataset contains approximately 87,053 high-quality constraint-text pairs.
%为了构建高质量的训练数据，我们对生成的数据从相关性、专业性、语言质量和安全合规性四个角度。进行了以下步骤。我们利用DeepSeek-R1[94]、Qwen2.5-72B-Instruct[93]进行独立评分，评分均值高于阈值0.75的样本被保留，构成约束-文本对的训练样本。最终构建的训练数据集包含约87,053条高质量的约束-文本对。

\subsection{Constrained Generative Model Training}
This paper selects LLaMa3-8B and GPT-2 as base models for fine-tuning to create constrained generation models. 
LLaMa3-8B, as a representative of the latest generation of open-source large language models, possesses powerful language understanding and generation capabilities, 
making it particularly suitable for complex constrained generation tasks requiring deep semantic understanding; while GPT-2, as a relatively lightweight but stable performance 
model, has significantly lower training and inference resource requirements, making it more suitable for practical applications under conditions of limited computational resources.

First, a topic type recognition module is introduced, which captures the category feature representation of the text through a multi-head attention mechanism, 
as shown in Equation~\eqref{eq:topic}:
\begin{equation}
H_{topic} = MultiHead(Q_t, K_t, V_t) 
\label{eq:topic}
\end{equation}
where $Q_t$, $K_t$, and $V_t$ represent the topic-related input, key, and value matrices, respectively.

In our output layer design, we adopt a structured decoding strategy to ensure format consistency and content integrity of the generated constraints 
through a conditional constraint generation mechanism, as shown in Equation~\eqref{eq:constraint}:
\begin{equation}
P(c_t|c_{<t},x,l) = softmax(W_o \cdot h_t + b_o)
\label{eq:constraint}
\end{equation}

where $c_t$ is the currently generated constraint text segment (such as constraint components like "topic", "key points", or "avoid" and their corresponding values), 
$c_{<t}$ is the already generated constraint sequence, $x$ is the input text, and $l$ is the topic type label.

The model training employs a two-stage training method with gradient accumulation. The first stage focuses on developing topic type recognition capability, 
using a supervised learning paradigm to enable the model to accurately judge the category attributes of input text; the second stage conducts fine-grained training 
for constraint generation based on recognition results, strengthening the model's ability to generate corresponding constraints according to different topic types. 
The training parameters are shown in Table~\ref{tab:parameters}:
\begin{table}[h]
\centering
\caption{Fine-tuning Parameters}
\label{tab:parameters}
\begin{tabular}{ccc}
\toprule
Parameter & LLaMa3-8B & GPT-2 \\
\midrule
learning\_rate & 1e-5 & 1e-4 \\
batch\_size & 32 & 64 \\
epoch & 3 & 6 \\
weight\_decay & 1e-3 & 1e-3 \\
\bottomrule
\end{tabular}
\end{table}

\section{Toxicity Reduction  Experiments}\label{sec:experiments}
\subsection{Experimental Setup}\label{subsubsec:setup}

\begin{table*}[t]
\caption{Experimental results comparing different methods for toxicity control.}
\label{tab:toxicity-results}
\centering
\begin{tabular}{llccccccc}
\toprule
 & & \multicolumn{3}{c}{Non-Toxic Scenario} & \multicolumn{3}{c}{Toxic Scenario} \\
\cmidrule(lr){3-5} \cmidrule(lr){6-8}
Category & Method & Max. Tox.$\downarrow$ & Tox. Prob.$\downarrow$ & PPL$\downarrow$ & Max. Tox.$\downarrow$ & Tox. Prob.$\downarrow$ & PPL$\downarrow$ \\
\midrule
Base Model & GPT-2 & 0.457 & 38.20\% & \textbf{11.29} & 0.759 & 84.20\% & \textbf{11.85} \\
\midrule
\multirow{2}{*}{Decoding-based} & DEXPERTS & \textbf{0.292} & \textbf{10.00\%} & 32.55 & \underline{0.492} & \underline{42.20\%} & \underline{33.59} \\
 & GeDi & 0.387 & 24.80\% & 38.21 & \textbf{0.430} & \textbf{34.20\%} & 47.42 \\
\midrule
\multirow{4}{*}{Prompt-based} & SD ($\lambda$=10) & 0.424 & 32.30\% & 13.02 & 0.723 & 80.60\% & 14.21 \\
 & SD ($\lambda$=50) & 0.373 & 23.10\% & 18.80 & 0.649 & 80.90\% & 19.66 \\
 & SD ($\lambda$=100) & 0.355 & 20.30\% & 21.09 & 0.623 & 65.50\% & 23.32 \\
 & REI & 0.496 & 40.60\% & 30.56 & 0.711 & 85.00\% & 31.20 \\
\midrule
\multirow{4}{*}{Fine-tuning-based} & DAPT & \underline{0.331} & \underline{18.90\%} & 19.72 & 0.558 & 57.00\% & 22.47 \\
 & ATCON & 0.482 & 42.00\% & 62.95 & 0.746 & 85.10\% & 69.51 \\
 & InstructionCTG & 0.360 & 38.60\% & 23.90 & 0.694 & 65.70\% & 27.56 \\
 & Ours & 0.304 & 21.20\% & \underline{12.31} & 0.608 & 61.90\% & \underline{14.01} \\
\bottomrule
\end{tabular}
\vspace{-2pt}
\end{table*}

\begin{figure*}[t]
\centering
\includegraphics[width=1\textwidth]{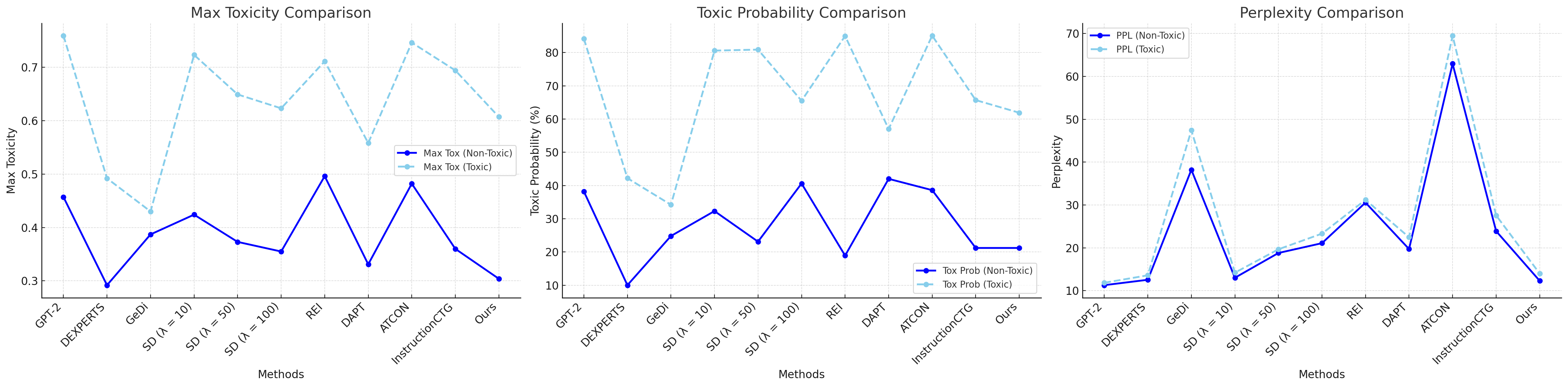}
\caption{Comparison of toxicity control performance across different methods. The bars show three metrics: Maximum Toxicity ($\downarrow$), Toxicity Probability ($\downarrow$), and Perplexity ($\downarrow$). }
\label{fig:toxicity-comparison}
\end{figure*}

Our experiments utilize GPT2-large\cite{radford2019language} as the base model to evaluate the effectiveness of various control strategies. For comprehensive assessment, we conduct experiments on the RealToxicityPrompts (RTP)\cite{gehman2020realtoxicityprompts} dataset, which comprises 100,000 text segments extracted from online English content. Each segment's first half serves as a continuation prompt, annotated with toxicity scores measured through the Perspective API~\cite{hosseini2017deceiving}.

From this dataset, we randomly sampled 10,000 prompts and, after filtering instances lacking toxicity score annotations, obtained 9,907 valid prompts. These were further categorized into two experimental scenarios:
\begin{itemize}
\item \textbf{Non-toxic Scenario}: 7,785 prompts with toxicity scores below 0.5
\item \textbf{Toxic Scenario}: 2,122 prompts with toxicity scores above 0.5
\end{itemize}

For evaluation purposes, the model generates continuations ranging from 5 to 20 tokens for each prompt.

\subsection{Evaluation Methodology}\label{subsubsec:evaluation}

Following Leong et al.\cite{leong-etal-2023-self}, we employ a fine-tuned DeBERTa-v3-large\cite{he2023debertav} model trained to approximate the Perspective API's toxicity probability through KL-divergence minimization. This model demonstrates robust performance with 94.87\% accuracy and 98.54\% AUROC on a 10,000-sample subset, indicating its effectiveness as a reliable alternative to the Perspective API for toxicity assessment while avoiding API rate limitations.

\subsection{Baseline Methods}\label{subsubsec:baselines}

We compare our approach against several baseline methods, all implemented using GPT2-large as the foundation model:

\paragraph{Fine-tuning-based Methods}
\begin{itemize}
\item \textbf{DAPT}\cite{gururangan-etal-2020-dont}: Continues pre-training on domain-specific unlabeled data to adapt the model to target domain characteristics
\item \textbf{ATCON}\cite{keskar2019ctrl}: Incorporates additional control-related tasks during pre-training and utilizes special control codes for style-specific generation
\item \textbf{InstructionCTG}\cite{zhou2023controlled}: Achieves effective control through weakly supervised data synthesis, natural language templates, and instruction tuning
\end{itemize}

\paragraph{Decoding-based Methods}
\begin{itemize}
\item \textbf{GeDi}~\cite{krause-etal-2021-gedi-generative}: Employs a discriminator model to guide generation by incorporating prediction scores into token probability distribution using Bayes' rule
\item \textbf{DEXPERTS}~\cite{liu-etal-2021-dexperts}: Utilizes expert and anti-expert models, combining their token predictions through weighted probability aggregation
\end{itemize}

\paragraph{Prompt-based Methods}
\begin{itemize}
\item \textbf{SD}~\cite{schick2021self}: Guides generation through specific prompt templates, comparing outputs with and without prompts to filter undesired content
\item \textbf{REI}~\cite{zheng2023toward}: Introduces regular expression instruction methodology, combining specific markup language with few-shot learning and fine-tuning strategies
\end{itemize}

For all methods, we employ nucleus sampling~\cite{Holtzman2020The} with $p = 0.9$, generating 25 continuations per prompt.

\subsection{ Results and Analysis}

\subsubsection{Main Results}\label{subsec:analysis}

\begin{table*}[htbp]
    \centering
    \caption{Comparison of toxicity metrics under different contexts}
    \label{tab:toxicity_comparison}
    \begin{subtable}{0.48\textwidth}
        \centering
        \caption{Results under non-toxic context}
        \label{tab:nontoxic_results}
        \begin{tabular}{lccccc}
            \toprule
            Model & Severe Tox & Sex & Threat & Profanity  & Id. Attack \\
            \midrule
            GPT-2 & 10.04 & 18.76 & 5.87 & 41.50 & 5.45 \\
            DEXPERTS & 6.95 & 9.35 & 5.65 & 18.77 & 5.76 \\
            GeDi & 4.21 & 13.41 & 3.92 & 12.78 & 5.53 \\
            SD($\lambda$ = 50) & \underline{1.75} & 12.88 & \underline{0.78} & 10.53 & \underline{1.87} \\
            REI & 5.37 & 16.17 & 1.59 & \underline{5.64} & \textbf{1.15} \\
            DAPT & 2.54 & \underline{9.63} & 3.77 & \textbf{5.23} & 2.63 \\
            ATCON & 2.58 & 11.64 & 1.84 & 21.58 & 2.17 \\
            InstructionCTG & 2.39 & 9.88 & 1.93 & 19.87 & 2.54 \\
            Ours & \textbf{1.80} & \textbf{8.24} & \textbf{1.56} & 7.16 & 2.32 \\
            \bottomrule
        \end{tabular}
    \end{subtable}
    \hfill
    \begin{subtable}{0.48\textwidth}
        \centering
        \caption{Results under toxic context}
        \label{tab:toxic_results}
        \begin{tabular}{lccccc}
            \toprule
            Model & Severe Tox & Sex & Threat & Profanity & Id. Attack \\
            \midrule
            GPT-2 & 65.81 & 49.52 & 23.84 & 68.47 & 36.80 \\
            DEXPERTS & 29.04 & 16.42 & 15.83 & 24.50 & 14.22 \\
            GeDi & 14.85 & 14.58 & 25.81 & 19.61 & 15.41 \\
            SD ($\lambda$ = 50) & 24.71 & 21.52 & 11.25 & 21.15 & 19.79 \\
            REI & 26.58 & 29.11 & 24.32 & 38.42 & 11.53 \\
            DAPT & \underline{13.47} & \underline{16.77} & \underline{4.26} & \underline{18.50} & \underline{9.54} \\
            ATCON & 19.32 & 17.15 & 19.80 & 25.77 & 13.27 \\
            InstructionCTG & 22.91 & 19.56 & 8.50 & 24.64 & 11.36 \\
            Ours & \textbf{15.78} & \textbf{10.53} & \textbf{6.78} & \textbf{19.80} & \textbf{5.47} \\
            \bottomrule
        \end{tabular}
    \end{subtable}
\end{table*}

As demonstrated in Table~\ref{tab:toxicity-results}, our proposed method exhibits significant toxicity control capabilities while maintaining low perplexity in both toxic and non-toxic contexts. In non-toxic contexts, our approach demonstrates superior performance compared to other fine-tuning-based methods in controlling toxicity continuation, specifically achieving a maximum toxicity score (0.304) and toxicity probability (21.20\%), while maintaining exceptional language fluency (PPL = 12.31). Compared to prompt-learning methods, our approach demonstrates advantages across both toxicity control metrics and perplexity measures, highlighting its multi-dimensional performance benefits.

In toxic contexts, while decoding-based methods (such as DEXPERTS and GeDi) show certain advantages in automated evaluation metrics (e.g., DEXPERTS achieving a maximum toxicity score of 0.292 and toxicity probability of 10.00\%), these methods face significant computational complexity challenges. Specifically, decoding methods typically require deployment of larger model architectures and implementation of probability distribution rewriting mechanisms.

These requirements lead to increased computational costs and higher perplexity scores (e.g., GeDi's perplexity scores of 38.21 and 47.42 in non-toxic and toxic contexts, respectively), potentially significantly impacting the fluency and naturalness of generated text.

\subsubsection{Fine-grained Analysis of Toxicity Control}
Our method demonstrates excellent control across five toxic categories under both non-toxic and toxic contexts as shown in Table~\ref{tab:toxicity_comparison}. 

In non-toxic settings (Table~\ref{tab:nontoxic_results}), we achieve superior performance in controlling Severe Toxicity (1.80) and Sex-related content (8.24), significantly outperforming baseline methods. While our approach shows slightly higher Profanity scores (7.16) compared to REI (5.64) and DAPT (5.23), the overall performance remains competitive.

In toxic contexts (Table~\ref{tab:toxic_results}), our method maintains balanced performance across all categories with notably better control over Threat (6.78) and Identity Attack (5.47), resulting in the lowest overall risk profile among all compared methods. These comprehensive evaluations across different toxicity dimensions and environmental contexts demonstrate the effectiveness and stability of our proposed method in controlling various forms of toxic content generation.

\subsection{Validation on Other Large Language Models}

To further validate the scalability and effectiveness of our proposed method on large language models, we conducted comprehensive toxicity evaluations on the llama3-8b model under toxic contexts. The experiments utilized 2,000 toxic prompts from the RealToxicityPrompts (RTP) dataset. To provide a thorough assessment of model practicality, we incorporated additional metrics including average generation length and rejection rate.The generation hyperparameters and results are listed in Table~\ref{tab:generation_params} and Table ~\ref{tab:llama_comparison}.
\begin{table}[htbp]
    \centering
    \begin{minipage}[b]{0.48\textwidth}
        \centering
        \caption{Generation hyperparameters}
        \label{tab:generation_params}
        \begin{tabular}{lc}
            \toprule
            Parameter & Value \\
            \midrule
            temperature & 0.70 \\
            top-p & 0.90 \\
            top-k & 50 \\
            max-length & 50 \\
            \bottomrule
        \end{tabular}
    \end{minipage}%
    \hspace{0.04\textwidth} % 设置两个表格之间的间距
    \begin{minipage}[b]{0.48\textwidth}
        \centering
        \caption{Comparison of toxicity control capabilities between base model and our method}
        \label{tab:llama_comparison}
        \begin{tabular}{lcc}
            \toprule
            Metrics & llama3-8b & llama3-8b\_PE-CG \\
            \midrule
            Max. Tox. (Toxic) & 0.25 & \textbf{0.15} \\
            Tox. Prob. (Toxic) & 15\% & \textbf{7\%} \\
            PPL (Toxic) & \textbf{10.50} & 11.56 \\
            Length (avg tokens) & 14.90 & \textbf{25.70} \\
            Refuse & 12\% & \textbf{5\%} \\
            \bottomrule
        \end{tabular}
    \end{minipage}
\end{table}

\begin{table*}[t]
    \centering
    \caption{Definitions and criteria for complex words and difficult words}
    \label{tab:word_criteria}
    \begin{tabular}{l|p{0.35\textwidth}|p{0.35\textwidth}}
        \toprule
        Characteristic & Complex Words & Difficult Words \\
        \midrule
        Definition & Number of syllables in words & Words that are not in the Dale-Chall word list \\
        \midrule
        Criterion & Words with three or more syllables & Check if word exists in Dale-Chall word list \\
        \midrule
        Examples & organization, unbelievable & catalyst, jurisprudence \\
        \midrule
        Application & Evaluates linguistic complexity of text & Evaluates vocabulary difficulty of text \\
        \bottomrule
    \end{tabular}
\end{table*}
In our experimental setup, while the base llama3-8b model demonstrated inherent safety mechanisms, there remained significant room for improvement across multiple metrics.

With constraint guidance, maximum toxicity reduced from 0.25 to 0.15 and toxicity probability decreased from 15\% to 7\%. Average generation length increased from 14.90 to 25.70 tokens, while the rejection rate for toxic prompts decreased from 9.12\% to 8.05\%.

These results empirically validate the effectiveness of our approach, showing that it not only improves security controls but also reduces rejection rates, ensuring the usefulness of the model.

\section{Readability-Controlled Summarization}

\subsection{Experimental Setup}

To evaluate the balance between controllability and utility of our proposed method, we follow the experimental protocol of ~\cite{wang-demberg-2024-rsa} and utilize LLaMA-2-7B-chat\cite{touvron2023llama} as the base model. The evaluation is conducted on the CNN/DailyMail\cite{see-etal-2017-get} dataset, comprising 11,490 news articles, to assess the model's capability in generating text with varying knowledge depths and tones.

The evaluation methodology systematically examines the model's ability to adapt to different audience comprehension levels by prefixing each article with specific instructions, such as:
\begin{itemize}
    \item Summarize the following news article for a primary-school student.
    \item Summarize the following news article for a college professor.
\end{itemize}

\subsection{Baseline Methods}
We compare our approach with three baseline methods:

\begin{enumerate}
    \item \textbf{Direct Output}: Direct generation using the base LLaMA-2-7B-chat model.
    
    \item \textbf{Style Transfer}~\cite{styleformer2021}: Post-processing the output through a style transformation model to adjust text formality).
    
    \item \textbf{CNN/DailyMail Fine-tuning}: Fine-tuning on the CNN/DailyMail dataset using two techniques:
    \begin{itemize}
        \item \textbf{Dynamic Word Unit Prediction} ~\cite{cao-wang-2021-inference}: Encodes readability levels (e.g., Flesch-Kincaid scores) as auxiliary features for the decoder, enabling dynamic word unit prediction during the decoding phase for direct style control during inference without model retraining.
        
        \item \textbf{Controllable Readability} ~\cite{ribeiro-etal-2023-generating}: Assigns readability scores to each text entry through manual annotation, utilizing model-evaluated readability scores of summaries in the CNN/DailyMail dataset as labels.
    \end{itemize}
\end{enumerate}

\subsection{Evaluation Metrics}
We evaluate the model’s controllability from both readability and summary quality. 
\subsubsection{Readability Metrics}
The readability assessment employs four widely-used metrics.

\begin{itemize}
    \item \textbf{Flesch Reading Ease (FRE)}: Evaluates text readability through word length and sentence length analysis.
    
    \item \textbf{Dale-Chall Readability (DCR)}: Considers vocabulary difficulty and sentence length, focusing on the usage of difficult words.
    
    \item \textbf{Gunning Fog Index (GFI)}: Emphasizes educational level requirements by examining sentence length and complex word proportions.
    
    \item \textbf{Coleman-Liau Index (CLI)}: Estimates required education level based on character, word, and sentence statistical features.
\end{itemize}

The definitions and evaluation criteria for complex words and difficult words as shown in Table~\ref{tab:word_criteria}, which serve as key components in our readability assessment metrics.
\subsubsection{Quality Metrics}
For semantic evaluation, we utilized BERTScore (BS), which leverages BERT's contextual representations to measure semantic similarity between generated and reference texts. Additionally, we incorporated Rouge-L (RG-L) to evaluate content overlap through longest common subsequence calculation, thereby assessing content completeness and accuracy. 
\subsubsection{Comprehensive  Metrics}
To provide a more holistic assessment independent of reference texts, we introduced GPT-4 scoring as a comprehensive metric that evaluates language fluency, contextual consistency, and readability.

\begin{table}[h]
\centering
\small  % 缩小字体
\setlength{\tabcolsep}{3pt}  % 减少列间距
\caption{Comprehensive evaluation results comparing different methods across readability and quality metrics}
\label{tab:results}
\begin{tabular}{lcccc}
\toprule
\textbf{Method} & \textbf{FRE}$\uparrow$ & \textbf{DCR}$\downarrow$ & \textbf{GFI}$\downarrow$ & \textbf{CLI}$\downarrow$ \\
\midrule
Default (LLaMA-2-7B-chat) & 53.57 & 10.48 & 14.08 & 11.69 \\
Style Transfer & 70.79 & 8.51 & 11.02 & 8.13 \\
Dynamic Word Unit Prediction & {75.70} & 9.59 & {8.26} & {8.50} \\
Controllable Readability & \textbf{83.20} & \textbf{6.60} & \textbf{6.30} & \textbf{6.80} \\
Ours & \underline{79.58} & \underline{7.52} &  \underline{8.02} &  \underline{7.26} \\
\bottomrule
\end{tabular}
\vspace{8mm}
\begin{tabular}{lccc}
\toprule
\textbf{Method} & \textbf{BS}$\uparrow$ & \textbf{RG-L}$\uparrow$ & \textbf{GPT4}$\uparrow$ \\
\midrule
Default (LLaMA-2-7B-chat) & \textbf{87.33} & \underline{34.63} & 82.57 \\
Style Transfer & 85.87 & 27.68 & \underline{86.55} \\
Dynamic Word Unit Prediction & \underline{86.98} & \textbf{37.88} & 88.38 \\
Controllable Readability & 86.80 & 30.75 & 87.16 \\
Ours & 84.94 & 24.97 & \textbf{92.56} \\
\bottomrule
\end{tabular}
\end{table}
\subsection{Results and Analysis}

Our experimental results as presented in Table~\ref{tab:results}. The experimental results demonstrate that our method achieved substantial improvements, particularly in readability metrics, with a 26.01 increase in the FRE score compared to the baseline. While some existing methods show competitive performance in individual readability metrics, our approach distinguishes itself through its balanced performance across all dimensions. Notably, our method achieved a GPT-4 score of 92.56, significantly outperforming other approaches and demonstrating superior overall text quality.

Although existing approaches such as Controllable Readability demonstrate marginally superior performance in certain individual metrics, we identify three significant limitations in these methods. First, they focus exclusively on readability control as a single dimension, neglecting the crucial aspect of content integrity. Second, these methods employ relatively coarse-grained control mechanisms that preclude precise adjustments of text characteristics. Third, their training on specific datasets results in limited generalization capability, restricting their effectiveness across diverse application scenarios.

This comprehensive performance is due to dynamic constraint guidance, which enables precise readability control while maintaining content integrity.

\section{Conclusion}
This paper proposes a controllable text generation method based on constraint guidance, which decomposes the task into three key steps: prediction type, constraint construction, and guided generation. By constructing a dataset containing 90,000 text-constraint pairs, precise control of different topic types is achieved. Experiments on the LLaMa3-8B and GPT-2 models verify the effectiveness of this method in toxicity control and readability adjustment tasks, which not only reduces harmful outputs, but also maintains a low perplexity, while achieving a better balance between readability and content quality.

\bibliography{reference}

\begin{thebibliography}{10}
\providecommand{\url}[1]{#1}
\csname url@samestyle\endcsname
\providecommand{\newblock}{\relax}
\providecommand{\bibinfo}[2]{#2}
\providecommand{\BIBentrySTDinterwordspacing}{\spaceskip=0pt\relax}
\providecommand{\BIBentryALTinterwordstretchfactor}{4}
\providecommand{\BIBentryALTinterwordspacing}{\spaceskip=\fontdimen2\font plus
\BIBentryALTinterwordstretchfactor\fontdimen3\font minus \fontdimen4\font\relax}
\providecommand{\BIBforeignlanguage}[2]{{%
\expandafter\ifx\csname l@#1\endcsname\relax
\typeout{** WARNING: IEEEtran.bst: No hyphenation pattern has been}%
\typeout{** loaded for the language `#1'. Using the pattern for}%
\typeout{** the default language instead.}%
\else
\language=\csname l@#1\endcsname
\fi
#2}}
\providecommand{\BIBdecl}{\relax}
\BIBdecl

\bibitem{touvron2023llama}
H.~Touvron, L.~Martin, and K.~e.~a. Stone, ``Llama 2: Open foundation and fine-tuned chat models,'' \emph{arXiv preprint arXiv:2307.09288}, 2023.

\bibitem{achiam2023gpt}
J.~Achiam, S.~Adler, and S.~e.~a. Agarwal, ``Gpt-4 technical report,'' \emph{arXiv preprint arXiv:2303.08774}, 2023.

\bibitem{shirish2019ctrl}
N.~S. Keskar, B.~McCann, and L.~R. e.~a. Varshney, ``Ctrl: a conditional transformer language model for controllable generation,'' \emph{arXiv e-prints}, 2019.

\bibitem{pascual-etal-2021-plug-play}
D.~Pascual, B.~Egressy, and C.~e.~a. Meister, ``A plug-and-play method for controlled text generation,'' in \emph{Findings of the Association for Computational Linguistics: EMNLP 2021}, 2021, pp. 3973--3997.

\bibitem{qian-etal-2022-controllable}
J.~Qian, L.~Dong, and Y.~e.~a. Shen, ``Controllable natural language generation with contrastive prefixes,'' in \emph{Findings of the Association for Computational Linguistics: ACL 2022}, 2022, pp. 2912--2924.

\bibitem{houlsby2019parameter}
N.~Houlsby, A.~Giurgiu, and S.~e.~a. Jastrzebski, ``Parameter-efficient transfer learning for nlp,'' in \emph{International Conference on Machine Learning}, 2019, pp. 2790--2799.

\bibitem{ouyang2022training}
L.~Ouyang, J.~Wu, and X.~e.~a. Jiang, ``Training language models to follow instructions with human feedback,'' \emph{Advances in Neural Information Processing Systems}, pp. 27\,730--27\,744, 2022.

\bibitem{gu-etal-2022-distributional}
Y.~Gu, X.~Feng, and S.~e.~a. Ma, ``A distributional lens for multi-aspect controllable text generation,'' in \emph{Proceedings of the 2022 Conference on Empirical Methods in Natural Language Processing}, 2022, pp. 1023--1043.

\bibitem{liu2024what}
W.~Liu, W.~Zeng, and K.~e.~a. He, ``What makes good data for alignment? a comprehensive study of automatic data selection in instruction tuning,'' in \emph{The Twelfth International Conference on Learning Representations}, 2024.

\bibitem{shin2020autoprompt}
T.~Shin, Y.~Razeghi, and R.~L. e.~a. Logan~IV, ``Autoprompt: Eliciting knowledge from language models with automatically generated prompts,'' \emph{arXiv preprint arXiv:2010.15980}, 2020.

\bibitem{pathania2024enhancing}
K.~Pathania, ``Enhancing conditional image generation with explainable latent space manipulation,'' \emph{arXiv preprint arXiv:2408.16232}, 2024.

\bibitem{krause-etal-2021-gedi-generative}
B.~Krause, A.~D. Gotmare, and B.~e.~a. McCann, ``{G}e{D}i: Generative discriminator guided sequence generation,'' in \emph{Findings of the Association for Computational Linguistics: EMNLP 2021}, 2021, pp. 4929--4952.

\bibitem{peng2023instruction}
B.~Peng, C.~Li, and P.~e.~a. He, ``Instruction tuning with gpt-4,'' \emph{arXiv preprint arXiv:2304.03277}, 2023.

\bibitem{chen2024sharegpt4v}
L.~Chen, J.~Li, and X.~e.~a. Dong, ``Sharegpt4v: Improving large multi-modal models with better captions,'' in \emph{European Conference on Computer Vision}, 2024, pp. 370--387.

\bibitem{liu2023chatcounselor}
J.~M. Liu, D.~Li, and H.~e.~a. Cao, ``Chatcounselor: A large language models for mental health support,'' \emph{arXiv preprint arXiv:2309.15461}, 2023.

\bibitem{zeng2020meddialog}
G.~Zeng, W.~Yang, and Z.~e.~a. Ju, ``Meddialog: Large-scale medical dialogue datasets,'' in \emph{Proceedings of the 2020 Conference on Empirical Methods in Natural Language Processing (EMNLP)}, 2020, pp. 9241--9250.

\bibitem{yue2023disc}
S.~Yue, W.~Chen, and S.~e.~a. Wang, ``Disc-lawllm: Fine-tuning large language models for intelligent legal services,'' \emph{arXiv preprint arXiv:2309.11325}, 2023.

\bibitem{huang2023lawyer}
Q.~Huang, M.~Tao, and C.~e.~a. Zhang, ``Lawyer llama technical report,'' \emph{arXiv preprint arXiv:2305.15062}, 2023.

\bibitem{bhatia2024fintral}
G.~Bhatia, E.~M.~B. Nagoudi, and H.~e.~a. Cavusoglu, ``Fintral: A family of gpt-4 level multimodal financial large language models,'' \emph{arXiv preprint arXiv:2402.10986}, 2024.

\bibitem{ji2023beavertails}
J.~Ji, M.~Liu, and J.~e.~a. Dai, ``Beavertails: Towards improved safety alignment of llm via a human-preference dataset,'' \emph{Advances in Neural Information Processing Systems}, vol.~36, pp. 24\,678--24\,704, 2023.

\bibitem{huang2023trustgpt}
Y.~Huang, Q.~Zhang, and L.~e.~a. Sun, ``Trustgpt: A benchmark for trustworthy and responsible large language models,'' \emph{arXiv preprint arXiv:2306.11507}, 2023.

\bibitem{radford2019language}
A.~Radford, J.~Wu, and R.~e.~a. Child, ``Language models are unsupervised multitask learners,'' \emph{OpenAI Blog}, p.~9, 2019.

\bibitem{gehman2020realtoxicityprompts}
S.~Gehman, S.~Gururangan, and M.~e.~a. Sap, ``Realtoxicityprompts: Evaluating neural toxic degeneration in language models,'' \emph{arXiv preprint arXiv:2009.11462}, 2020.

\bibitem{hosseini2017deceiving}
H.~Hosseini, S.~Kannan, and B.~e.~a. Zhang, ``Deceiving google's perspective api built for detecting toxic comments,'' \emph{arXiv preprint arXiv:1702.08138}, 2017.

\bibitem{leong-etal-2023-self}
C.~T. Leong, Y.~Cheng, and J.~e.~a. Wang, ``Self-detoxifying language models via toxification reversal,'' in \emph{Proceedings of the 2023 Conference on Empirical Methods in Natural Language Processing}, 2023, pp. 4433--4449.

\bibitem{he2023debertav}
P.~He, J.~Gao, and W.~Chen, ``De{BERT}av3: Improving de{BERT}a using {ELECTRA}-style pre-training with gradient-disentangled embedding sharing,'' in \emph{The Eleventh International Conference on Learning Representations}, 2023.

\bibitem{gururangan-etal-2020-dont}
S.~Gururangan, A.~Marasovi{\'c}, and S.~e.~a. Swayamdipta, ``Don't stop pretraining: Adapt language models to domains and tasks,'' in \emph{Proceedings of the 58th Annual Meeting of the Association for Computational Linguistics}, 2020, pp. 8342--8360.

\bibitem{keskar2019ctrl}
N.~S. Keskar, B.~McCann, and L.~R. e.~a. Varshney, ``Ctrl: A conditional transformer language model for controllable generation,'' \emph{arXiv preprint arXiv:1909.05858}, 2019.

\bibitem{zhou2023controlled}
W.~Zhou, Y.~E. Jiang, and E.~e.~a. Wilcox, ``Controlled text generation with natural language instructions,'' in \emph{International Conference on Machine Learning}, 2023, pp. 42\,602--42\,613.

\bibitem{liu-etal-2021-dexperts}
A.~Liu, M.~Sap, and X.~e.~a. Lu, ``{DE}xperts: Decoding-time controlled text generation with experts and anti-experts,'' in \emph{Proceedings of the 59th Annual Meeting of the Association for Computational Linguistics and the 11th International Joint Conference on Natural Language Processing (Volume 1: Long Papers)}, 2021, pp. 6691--6706.

\bibitem{schick2021self}
T.~Schick, S.~Udupa, and H.~Sch{\"u}tze, ``Self-diagnosis and self-debiasing: A proposal for reducing corpus-based bias in nlp,'' \emph{Transactions of the Association for Computational Linguistics}, pp. 1408--1424, 2021.

\bibitem{zheng2023toward}
X.~Zheng, H.~Lin, and X.~e.~a. Han, ``Toward unified controllable text generation via regular expression instruction,'' \emph{arXiv preprint arXiv:2309.10447}, 2023.

\bibitem{Holtzman2020The}
A.~Holtzman, J.~Buys, and L.~e.~a. Du, ``The curious case of neural text degeneration,'' in \emph{International Conference on Learning Representations}, 2020.

\bibitem{wang-demberg-2024-rsa}
Y.~Wang and V.~Demberg, ``{RSA}-control: A pragmatics-grounded lightweight controllable text generation framework,'' in \emph{Proceedings of the 2024 Conference on Empirical Methods in Natural Language Processing}, 2024, pp. 5561--5582.

\bibitem{see-etal-2017-get}
A.~See, P.~J. Liu, and C.~D. Manning, ``Get to the point: Summarization with pointer-generator networks,'' in \emph{Proceedings of the 55th Annual Meeting of the Association for Computational Linguistics (Volume 1: Long Papers)}, 2017, pp. 1073--1083.

\bibitem{styleformer2021}
\BIBentryALTinterwordspacing
P.~Damodaran, ``Styleformer,'' 2021. [Online]. Available: \url{https://github.com/PrithivirajDamodaran/Styleformer}
\BIBentrySTDinterwordspacing

\bibitem{cao-wang-2021-inference}
S.~Cao and L.~Wang, ``Inference time style control for summarization,'' in \emph{Proceedings of the 2021 Conference of the North American Chapter of the Association for Computational Linguistics: Human Language Technologies}, 2021, pp. 5942--5953.

\bibitem{ribeiro-etal-2023-generating}
L.~F.~R. Ribeiro, M.~Bansal, and M.~Dreyer, ``Generating summaries with controllable readability levels,'' in \emph{Proceedings of the 2023 Conference on Empirical Methods in Natural Language Processing}, 2023, pp. 11\,669--11\,687.

\end{thebibliography}

\end{document}